\documentclass{article}

\PassOptionsToPackage{numbers, compress}{natbib}

\usepackage[position,final]{neurips_2026}

\usepackage[utf8]{inputenc} %
\usepackage[T1]{fontenc}    %
\usepackage{hyperref}       %
\usepackage{url}            %
\usepackage{booktabs}       %
\usepackage{amsfonts}       %
\usepackage{nicefrac}       %
\usepackage{microtype}      %
\usepackage{xcolor}         %

\usepackage{graphicx}
\usepackage{subcaption}
\usepackage{latexsym}
\usepackage{enumitem}

\usepackage{inconsolata}

\usepackage{graphicx}
\usepackage{array}
\usepackage{booktabs}
\usepackage{tabularx}
\usepackage{amsmath}
\usepackage[most]{tcolorbox}
\newtcolorbox{miscalbox}[1][]{
  colback=blue!5, colframe=blue!40!black,
  fonttitle=\bfseries\small, title=#1,
  boxrule=0.4pt, top=2pt, bottom=2pt, left=4pt, right=4pt
}

\usepackage{xcolor}
\usepackage{amsmath}
\usepackage{amssymb}
\usepackage{mathtools}
\usepackage{amsthm}

\newcommand{\rparagraph}[1]{\vspace{0mm}\noindent\textbf{#1}}

\usepackage[capitalize,noabbrev]{cleveref}

\definecolor{miscalRed}{RGB}{255,155,133}      %
\definecolor{miscalOrange}{RGB}{244,197,66}    %
\definecolor{miscalBlue}{RGB}{74,124,156}      %
\definecolor{miscalTeal}{RGB}{93,190,170}      %

\definecolor{miscalCoralInk}{HTML}{B94D38}
\definecolor{miscalGoldInk}{HTML}{966E13}
\definecolor{miscalBlueInk}{HTML}{2E729D}
\definecolor{miscalTealInk}{HTML}{207D70}

\tcbset{
  miscal definition/.style={
    enhanced,
    colback=white,
    colframe=#1!45!white,
    coltitle=#1,
    boxrule=0.55pt,
    arc=2pt,
    outer arc=2pt,
    boxsep=0pt,
    left=10pt,
    right=10pt,
    top=9pt,
    bottom=8pt,
    before skip=10pt,
    after skip=8pt,
    fonttitle=\sffamily\bfseries\small,
    fontupper=\normalfont\normalsize,
    attach boxed title to top left={xshift=7pt,yshift*=-\tcboxedtitleheight/2},
    boxed title style={
      colback=white,
      colframe=white,
      boxrule=0pt,
      sharp corners,
      boxsep=0pt,
      left=4pt,
      right=4pt,
      top=1pt,
      bottom=1pt
    }
  }
}

\newtcolorbox{miscalRedBox}[1][]{miscal definition=miscalCoralInk,title={#1}}
\newtcolorbox{miscalOrangeBox}[1][]{miscal definition=miscalGoldInk,title={#1}}
\newtcolorbox{miscalBlueBox}[1][]{miscal definition=miscalBlueInk,title={#1}}
\newtcolorbox{miscalTealBox}[1][]{miscal definition=miscalTealInk,title={#1}}

\usepackage{hyperref}
\usepackage{url}

\title{Large Language Model Failures from Hallucination to Homogenization Are Different Facets of Miscalibration}

\author{%
  \setcounter{footnote}{1}%
  \textbf{Tiancheng Hu}$^{1,}$\thanks{Email: \texttt{th656@cam.ac.uk}}\;,
  \textbf{Caiqi Zhang}$^{1}$,
  \setcounter{footnote}{0}%
  \textbf{Dirk Hovy}$^{2,}$\thanks{Equal contribution.}\;, \textbf{and}
  \textbf{Nigel Collier}$^{1,}$\footnotemark[1]
  \vspace{0.2cm} \\
  $^{1}$~University of Cambridge, \quad
  $^{2}$~Bocconi University
}

\begin{document}

\maketitle

\begin{abstract}

We argue that calibration should become a primary evaluation and optimization target for large language models (LLMs). Failures typically studied in isolation, from confident hallucinations and homogenized outputs to inappropriate refusals, can all be understood as different facets of miscalibration. We propose a unified framework distinguishing four types: \emph{probabilistic}, where generated samples fail to match a specified distribution; \emph{semantic}, where confidence does not reflect correctness; \emph{distributional}, where outputs underrepresent the range of valid responses; and \emph{metacognitive}, where models misjudge their own competence. The stakes are particularly high in agentic settings, where miscalibrated decisions can affect subsequent actions and the outcome of an entire task. We argue that calibration is essential for evaluating not only what models can do, but also how reliably they do it. We call for calibration metrics to become standard in benchmark reporting, for training methods that preserve uncertainty and diversity, and for interaction designs that communicate model confidence to downstream decision-makers.
\end{abstract}

\section{Introduction}
\label{sec:intro}
Ask GPT-4o for a random number between 1 and 10, and it may say ``7'' over 60\% of the time, mirroring a well-documented human cognitive bias known as the \emph{Blue-Seven Phenomenon}~\citep{west2025base}.
Ask it how many Rs are in ``strawberry,'' and it will confidently, incorrectly answer ``two.''
Ask it to write a story featuring a Black protagonist, and it will pepper the text with shibboleths like ``finna'' and ``chile,'' collapsing the diversity of Black American speech into a narrow caricature~\citep{rooein2025biased}.
Ask it what 2+3 equals, and it may spend a thousand tokens ``reasoning'' before answering 5~\citep{chen2024not}.
These failures are typically studied in isolation: probabilistic reasoning, factuality, bias, and efficiency, respectively. \textbf{We argue that these seemingly disparate failures are facets of miscalibration, the systematic misalignment between a model's confidence and reality. Calibration must therefore become a primary evaluation and optimization target.} As LLMs are increasingly deployed in high-stakes domains, from medical diagnosis and legal research to financial advising and educational assessment, the ability to trust their outputs becomes essential. Yet trust requires more than capability; it demands reliability, which in turn depends on calibration~\cite{steyvers2025large}.

\begin{figure}[tbp!]
    \centering
    \includegraphics[width=1\columnwidth]{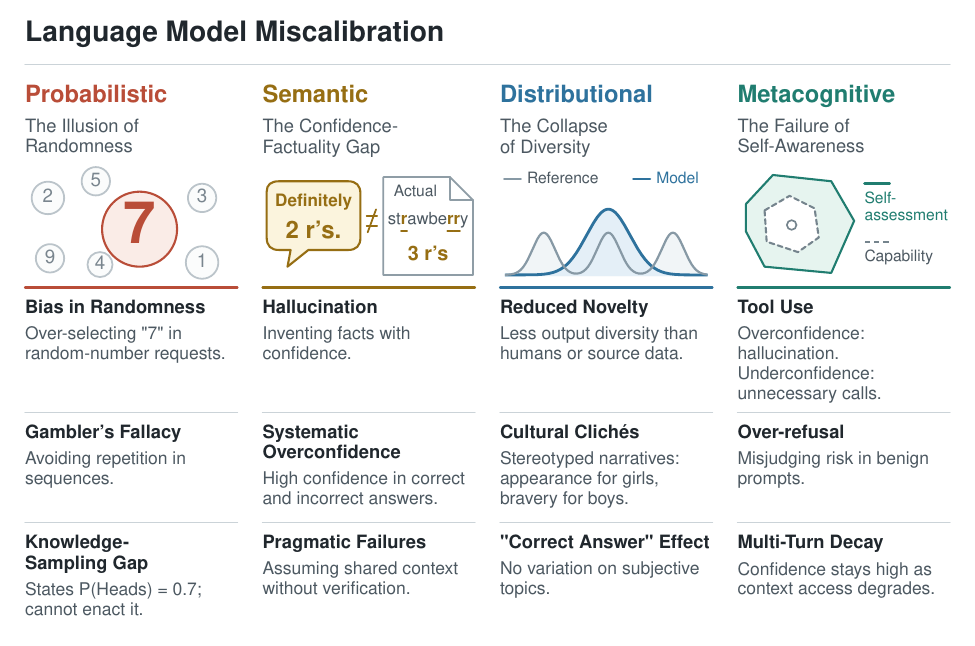}
    \caption{Four facets of LLM miscalibration. Each facet represents a distinct manifestation of the misalignment between a model's confidence and reality, from failures in probabilistic reasoning to breakdowns in self-awareness.}
    \label{fig:main}
\end{figure}

If miscalibration is the common thread, we need adequate tools to diagnose it. Current calibration evaluations of LLMs descend directly from classification metrics like Expected Calibration Error (ECE). While valuable, these
metrics are increasingly insufficient: LLMs are no longer merely classifiers but generators, simulators, reasoners, creators, and increasingly, autonomous agents. Their calibration failures extend far beyond what traditional metrics capture. This is particularly evident in agentic settings, where models repeatedly decide whether to act, gather evidence, verify results, or seek assistance. Each decision shapes subsequent observations and opportunities for recovery, so miscalibration can derail an entire trajectory. This paper therefore offers a unified framework for understanding these broader failures, organized around four key areas where miscalibration is both pervasive and under-appreciated:

\begin{enumerate}[nosep,leftmargin=*]
\item \textbf{Probabilistic Miscalibration: The Illusion of Randomness} (\S\ref{sec:probabilistic}). The model's failure to generate outputs that faithfully match a requested probability distribution. Example: sampling ``7'' 60\% of the time when asked for a random number between 1 and 10 (\emph{Blue-Seven}).  
\item \textbf{Semantic Miscalibration: The Confidence-Factuality Gap} (\S\ref{sec:semantic}). The gap between a model's expressed confidence and the factual or logical correctness of its statements. Example: confidently stating that ``strawberry'' contains 2 Rs (\emph{strawberry}).
    \item \textbf{Distributional Miscalibration: The Collapse of Diversity} (\S\ref{sec:distributional}). Collapsing the wide distribution over possible valid outputs into a narrow, high-probability set of clichés. Example: overusing shibboleths when prompted with a Black persona (\emph{finna}).
    \item \textbf{Metacognitive Miscalibration: The Failure of Self-Awareness} (\S\ref{sec:metacognitive}). The inability of an agentic model to accurately assess its own knowledge and decide when to act, defer, or seek help. Example: spending thousands of tokens ``thinking'' before answering (\emph{2+3}).
\end{enumerate}

Viewing these failures through a probabilistic
lens~\cite{shah-etal-2020-predictive} suggests a common
diagnostic question: \emph{what distribution does the
model produce, and what distribution should it match?}

\rparagraph{Contributions:}
\begin{enumerate}[nosep,leftmargin=*]
    \item A \textbf{unified framework} that links diverse model failures such as 
          hallucination, homogenization, over-refusal, and reasoning failures 
          (\S\ref{sec:framework}; see Figure~\ref{fig:main} for a visual overview).
\item A \textbf{diagnostic methodology} for analyzing these failures
    by comparing the distribution a model produces with the
    distribution it should match.
    \item A \textbf{call to action} for making calibration a standard
    part of model evaluation, training, and deployment with concrete recommendations for researchers, developers, and practitioners
    (\S\ref{sec:outlook}).
\end{enumerate}

\section{What Is Calibration?}
\label{sec:terminology}

\rparagraph{Pre-LLMs.} 
In traditional supervised learning, research has extensively covered uncertainty and calibration, with less emphasis on direct confidence estimation.
A model's \textit{confidence} is typically defined as the softmax probability assigned to a predicted class \citep{nixon2019measuring, corbiere2019addressing}. Because the output belongs to a fixed, enumerable set, extracting confidence is straightforward. 
Consequently, prior research has primarily focused on calibration methods to align predicted probabilities with empirical accuracy.
Prominent examples include Platt scaling \citep{platt1999probabilistic}, isotonic regression \citep{niculescu2005predicting}, and temperature scaling \citep{guo2017calibration}.

\rparagraph{Generative LLM Context.} 
The advent of generative LLMs necessitates a broader definition of calibration, as they produce free-form text lacking natural single-class probabilities. Confidence estimation in this setting is complicated by three primary factors. First, \emph{semantic equivalence} means that multiple distinct strings can represent the same underlying answer \citep{kuhnsemantic, zhang-etal-2024-luq}. Second, \emph{alignment distortions} arise because token probabilities are frequently skewed by decoding strategies and RLHF \citep{achiam2023gpt, tian2023just, kadavath2022language}. Finally, the \emph{intractable output space}, which grows exponentially with sequence length, precludes normalizing probabilities over all possible correct answers \citep{zhang-etal-2025-atomic}. These challenges motivate our work in Section~\ref{sec:framework}, where we propose a broader taxonomy of miscalibration that seeks to capture diverse failure modes.

\rparagraph{Measuring Calibration in Generative LLMs.}
These challenges do not make calibration measurement
intractable. Rather than evaluating probabilities over
individual token sequences, semantically equivalent
answers can be grouped together, while long-form
responses can be decomposed into factual claims
~\citep{kuhnsemantic,zhang-etal-2024-luq}.
Confidence can then be estimated through verbalized
probabilities, learned estimators, or repeated
sampling~\citep{tian2023just,manakul2023selfcheckgpt}.
For tasks with multiple valid responses, the sampled
output distribution can instead be compared with a
task-specific reference distribution. These approaches
avoid enumerating the full output space, although
their reliability still depends on how the prediction
target and reference distribution are defined.

\rparagraph{Confidence vs. Uncertainty.}
We distinguish two oft-conflated concepts: \textbf{confidence} estimates the probability that a \emph{specific answer} is correct (guiding acceptance and ranking)~\citep{zhang2026xconf}, while \textbf{uncertainty} measures the dispersion of beliefs over \emph{all possible answers} (guiding abstention and tool use) \citep{lin2023generating, zhang-etal-2024-luq}. Both can be miscalibrated; our framework addresses both. Formal definitions appear in \S\ref{app:notation}.

\rparagraph{Calibration vs. Alignment.}
Calibration and alignment are related but distinct concepts.
Alignment concerns whether a model's behavior follows
human goals or preferences, whereas calibration concerns
whether its predictions, confidence, and decisions track
their corresponding empirical targets. A model may follow
its intended behavioral policy while remaining overconfident
in incorrect answers. Conversely, a model may accurately
predict the consequences of an action without acting in
accordance with human preferences. Thus, improvements
in alignment do not necessarily imply improvements
in calibration, and vice versa.

\section{A Framework for LLM Miscalibration}
\label{sec:framework}
We propose viewing LLM failures through four facets of
miscalibration. These are not mutually exclusive and frequently co-occur in practice.
They differ in the model signal being evaluated and the
target it should track (Table~\ref{tab:facets}). Together, they are designed to span the space of confidence-reality mismatches in generative models. Our framework is diagnostic rather than  causal: we do not claim that miscalibration is the sole root cause of these failures, but that a wide range of systematic LLM failures can be productively analyzed as a miscalibration along one or more of these dimensions. We illustrate this with premature closure in clinical reasoning in Appendix~\ref{app:diagnostic}.

\begin{table*}[t]
\centering
\caption{The four facets of miscalibration differ in the
object being calibrated and the target it should track.
The facets are not mutually exclusive.}
\label{tab:facets}
\small
\setlength{\tabcolsep}{10pt}
\renewcommand{\arraystretch}{1.2}
\begin{tabularx}{\textwidth}{@{}
    >{\raggedright\arraybackslash}p{0.19\textwidth}
    >{\raggedright\arraybackslash}X
    >{\raggedright\arraybackslash}X
    @{}}
\toprule
\textbf{Facet} &
\textbf{Object being calibrated} &
\textbf{Target} \\
\midrule
\textbf{Probabilistic} &
Sampled outcomes &
An explicitly specified distribution \\
\textbf{Semantic} &
Confidence in a claim &
Empirical correctness \\
\textbf{Distributional} &
Distribution over valid outputs &
A reference distribution \\
\textbf{Metacognitive} &
Competence or action-value estimate &
Realized success or utility \\
\bottomrule
\end{tabularx}
\end{table*}

\subsection{Probabilistic Miscalibration: The Illusion of Randomness}
\label{sec:probabilistic}
\begin{miscalRedBox}[Probabilistic Miscalibration]
The model's output distribution $q_\theta(y \mid x)$ fails to match a target distribution $p^*(y \mid x)$ specified in the prompt.
\end{miscalRedBox}
\vspace{-1mm}

Calibration, at its core, is about matching probabilities to empirical frequencies. Here, the model fails to enact the very probabilities it can correctly describe.

\rparagraph{The Problem.} The model's generated distribution of outputs fails to match the target distribution $p^\star(y \mid x)$ described in the prompt. This can be conceptualized as a high divergence between the empirical distribution of samples from $q_\theta$ and the specified $p^\star$ (e.g., KL-divergence). This matters beyond simple queries: stochastic agent policies, best-of-N generation, and Monte Carlo uncertainty estimation all depend on accurate sampling. When the model cannot match a specified distribution, downstream systems inherit systematic bias.

\rparagraph{Random Number Generation.} Models cannot reliably produce random numbers. When prompted for a single uniformly random integer, aligned models display a strong preference for ``7,'' a well-documented psychological bias in humans \citep{doi:10.2466/pms.1971.33.2.373, west2025base}. Similarly, \citet{zhang2024forcing} found Llama-2-13B-chat assigns over 60\% of its probability mass to the number ``5'' when prompted for a random number between 1 and 10, and Mistral-7B-Instruct chooses the name ``Avery'' 40 times more often than we would expect based on the U.S. population.

\rparagraph{Sequential Sampling.} This miscalibration extends to sequences. Aligned models exhibit an aversion to repetition that mirrors the human gambler's fallacy, i.e., the belief that randomness means outcomes are conditional and alternate with some regularity (e.g., after three times red, the next roulette spin is bound to be black). However, a truly random process can contain long sequences of repeated outcomes. For aligned models, the most common outcome (the mode) is a sequence with \textit{zero} repetitions, a significant deviation from any true random sample~\citep{west2025base}.

\rparagraph{Knowledge-Behavior Gap.} Most tellingly, the previous behavior reveals a gap between a model's declarative knowledge and its sampling behavior. A model explicitly prompted with the ground truth of a biased coin (e.g., $P(Heads) = 0.7$) will accurately \textit{verbalize} this probability if asked, yet its internal log-probabilities used for sampling remain highly un-calibrated and fail to match the provided fact \citep{meister-etal-2025-benchmarking,kirchhof2025selfreflectllmscommunicateinternal}. The model can describe probability, but cannot enact it. This gap undermines any downstream system that relies on verbalized probabilities as a proxy for sampling behavior.

\subsection{Semantic Miscalibration: The Confidence-Factuality Gap}
\label{sec:semantic}
\begin{miscalOrangeBox}[Semantic Miscalibration]
The model's expressed confidence does not reflect the factual correctness of its claims.
\end{miscalOrangeBox}
\vspace{-1mm}

This lens moves beyond distribution shapes to examine the correctness of generated content. The core question: when a model sounds confident, should we believe it?

\rparagraph{The Problem.}
When faced with queries beyond its knowledge, a well-calibrated model should express uncertainty. Instead, current models often exhibit overconfidence in incorrect answers and underconfidence in correct ones~\citep{sung-etal-2025-grace}.

\rparagraph{Root Causes.}
The prevailing evaluation paradigms rely on binary right-or-wrong scoring, systematically rewarding guessing over expressing uncertainty \citep{kalai2025languagemodelshallucinate}. Even techniques like RLVR degrade calibration due to binary rewards~\citep{damani2025beyond}. This is compounded by a training dilemma: teaching "I don't know" risks models withholding knowledge they possess, while training only on correct answers forces guessing \citep{Schulman_2023_transcript}.

\textbf{Hallucination.} Hallucination, where models generate false or unsupported
content as if it were true, is a common manifestation
of semantic miscalibration. Despite significant improvements in model
capability, hallucination remains a persistent problem that many users have personal experiences of
~\cite{10.1162/COLI.a.16}. On SimpleQA, a benchmark of short factual questions, frontier models achieve accuracies around 50\%~\citep{wei2024simpleqa, OpenAI_GPT5_SystemCard_2025, gemini2.5}. This is not merely a capability limit: the original study finds that models consistently overstate their confidence, with stated confidence exceeding actual accuracy across all models tested~\citep{wei2024simpleqa}. This problem extends to long-form generation, where LLMs often fail to mark uncertain facts within extended outputs~\citep{yang-etal-2025-uncle, yang-etal-2025-logu, 10.5555/3692070.3692180}. Reinforcement-learning methods that encourage explicit uncertainty expressions can mitigate the problem, but they do not eliminate it~\citep{zhang2025lovec, 10.5555/3692070.3692180}. Overall, a large gap remains between what models know and how faithfully they communicate what they do not know.

\textbf{High-Stakes Domains.} Semantic miscalibration carries severe consequences in domains where errors cause direct harm. In healthcare, frontier models hallucinate in 23.4\% of medical reasoning tasks, yet express high confidence even in these  cases~\cite{kim2025medical}. This miscalibration compounds with data frequency: models hallucinate more on rare conditions with sparse training coverage~\citep{zhao2024wildhallucinations}, precisely where clinical expertise is most critical. Other high-stakes fields, such as the legal domain, face parallel risks, with models fabricating case citations and financial data while maintaining unwarranted confidence~\citep{10.1093/jla/laae003,https://doi.org/10.1111/jels.12413}.

\textbf{Presumptive Grounding.} Semantic miscalibration extends to pragmatic failures. LLMs frequently assume shared context that does not exist, being three times less likely than humans to seek clarification~\citep{shaikh-etal-2025-navigating}. This reflects miscalibration of the model's theory of mind: it overestimates the probability that its interpretation of user intent is correct. The extreme case is sycophancy, where models abandon their own knowledge to agree with incorrect user premises, miscalibrating their confidence in the user's correctness over their own~\citep{sharma2024towards}.

\subsection{Distributional Miscalibration: The Collapse of Diversity}
\label{sec:distributional}
\begin{miscalBlueBox}[Distributional Miscalibration]
The model collapses the space of valid outputs into a narrow, stereotypical mode.
\end{miscalBlueBox}
\vspace{-1mm}

While semantic miscalibration focuses on the correctness of a single output $\hat{y}$, distributional miscalibration concerns the properties of the entire predictive distribution $q_\theta(y \mid x)$.

\rparagraph{The Problem.} 
Many common use cases for LLMs, from creative writing and summarization to product recommendation and advice-seeking, are inherently underspecified and subjective. Instead of having a single correct answer $y$, there exists a distribution over valid responses that reflects the diversity of human preferences and perspectives. Distributional miscalibration occurs when the model's output distribution $q_\theta(y \mid x)$ fails to accurately represent this diversity. Some prior work refers to this as ``generative monoculture'': a significant narrowing of output diversity relative to the variety present in the source data~\cite{wu2025generative}. A common failure mode is when $q_\theta(y \mid x)$ concentrates probability mass on the most common or prototypical responses while underrepresenting less frequent but equally valid alternatives. In other words, the model captures the central tendency of acceptable responses but fails to represent their full range.

\textbf{Reduced Novelty.} In the domain of language and creativity, miscalibration manifests as a collapse in lexical and semantic diversity. For example, on NoveltyBench, current state-of-the-art LLMs produce significantly fewer unique and novel outputs compared to human writers \citep{zhang2025noveltybench}. This finding is corroborated by large-scale analyses of ``epistemic diversity," which demonstrate that LLMs generate a narrower range of real-world claims than a simple web search \citep{wright2025epistemic}. Crucially, both lines of research converge on a somewhat counter-intuitive scaling property: larger models can exhibit less diversity than their smaller counterparts. Furthermore, \citet{zhang2025noveltybench} trace the root of this problem, showing that each successive stage of alignment, from supervised fine-tuning to preference optimization, progressively reduces a model's output diversity. This finding suggests the very processes intended to make models more useful are simultaneously making them more uniform.

\textbf{Clichés and Flattened Identity.}
When the distribution concerns people, this miscalibration hardens into cliché. Children's stories generated for non-Western contexts frequently fixate on orientalist tropes, erasing the diverse reality of modern cultures \citep{rooein2025biased}. Similarly, recent work identifies ``echoes'' in narrative generation, quantifying a severe lack of plot diversity compared to humans \citep{doi:10.1073/pnas.2504966122}. In recommendation and interaction, this tendency ``flattens'' identity groups by collapsing a wide spectrum of valid possibilities into a narrow, stereotypically-associated mode \citep{Wang2025}. For instance, when a user implicitly reveals their race as Black while asking for college recommendations, models disproportionately suggest Historically Black Colleges and Universities (HBCUs), failing to represent the vast range of institutions that would be considered by a human advisor \citep{kantharuban-etal-2025-stereotype}.

\rparagraph{Opinion Homogenization.} 
This failure extends to representing human subjectivity. Instruction-tuning improves a model's ability to predict consensus views but actively harms its ability to capture pluralistic, high-entropy disagreements~\citep{hu2025simbench}. This leads to the ``correct answer'' effect, where models respond to nuanced survey questions about politics and morality with near-zero variation, effectively hallucinating a consensus where none exists~\citep{Park2024}. Even when a query is ambiguous (e.g., ``What book should I read next?''), models default to a single interpretation—often aligned with a WEIRD context~\cite{Henrich_Heine_Norenzayan_2010}—rather than representing the diversity of possible user needs~\citep{10.1162/TACL.a.24}. For generative tasks with many valid answers, a well-calibrated model should distribute its probability mass across multiple good sequences rather than collapsing to a single mode~\citep{10.5555/3692070.3693952,flores-etal-2025-improving}.

\subsection{Metacognitive Miscalibration: The Failure of Self-Awareness}
\label{sec:metacognitive}
\begin{miscalTealBox}[Metacognitive Miscalibration]
The model's self-assessment of competence is misaligned with its actual capability.
\end{miscalTealBox}
\vspace{-1mm}

The previous forms of miscalibration concerned the final output content. Metacognitive miscalibration concerns the model's internal self-assessment: the alignment between its confidence in its own capabilities and its actual performance.

\rparagraph{The Problem.} A well-calibrated model should ``know what it knows'': accurately estimating when it has sufficient knowledge to answer directly versus when it should use a tool, defer to a human, or refuse to answer. Current models often fail at this self-assessment, leading to both overconfident errors (hallucinating when they should abstain) and underconfident inefficiencies (invoking tools for trivial tasks). This becomes particularly consequential in agentic settings, where models must
repeatedly assess their competence to decide whether to reason further, invoke tools, or take actions. A miscalibrated decision can propagate through subsequent steps, causing agents to pursue flawed plans or miss opportunities to correct earlier mistakes. Reliable
self-assessment is therefore critical for agentic systems, where errors can compound over long sequences of decisions~\cite{yona2026hallucinationsunderminetrustmetacognition}.

\rparagraph{Miscalibration in Tool Use.} 
The decision to consult an external tool versus relying on parametric memory is a quintessential metacognitive challenge for agentic LLMs: the model must estimate its own confidence in answering directly and compare it to the expected utility of external retrieval. A well-calibrated model's ``tool use decision boundary" should align with its actual ``knowledge boundary"~\cite{wang2025theoryagentstoolusedecisionmakers}.
Miscalibration leads to two costly failure modes. An overconfident model neglects tools even when necessary, particularly for tasks requiring real-time information or precise arithmetic~\cite{wang2025theoryagentstoolusedecisionmakers}. Conversely, an underconfident model engages in tool overuse, unnecessarily invoking external tools for tasks solvable with its own knowledge, incurring significant latency and computational overhead~\cite{qian-etal-2025-smart}. The accumulation of minor errors from tool interactions can then degrade overall task performance~\cite{NEURIPS2024_e4c61f57}. The problem is further compounded by fragile tool selection, where models are often swayed by superficial phrasing in tool descriptions rather than actual functionality~\cite{faghih2025gaming}. These issues collectively point to a core deficit in ``decision calibration": a misalignment between a model's confidence in a chosen action and the actual utility of that choice~\cite{liu-etal-2024-uncertainty}.

\rparagraph{Miscalibration in Safety Systems.} Deployed models make constant decisions about whether a prompt is safe to answer. Conceptually, this decision is governed by an internal, often implicit, estimate of the probability of harm, $p(\text{harm} | x)$. A pervasive failure is over-refusal (also termed ``overkill" or ``false refusal"), where this probability is miscalibrated too high, causing the model to be overconfident that a benign prompt is harmful~\cite{pmlr-v267-cui25a,zhang2025falsereject}. This brittleness reveals a poorly calibrated understanding of risk, which often stems from a superficial analysis that relies on verbal shortcuts, becoming overly sensitive to specific keywords (e.g., `kill') regardless of the safe context~\cite{shi-etal-2024-navigating}. The challenge is further compounded by the need for user-specific safety, as the true $p(\text{harm})$ is conditioned on the user's profile: a query that is safe for the general population may be harmful to a vulnerable individual, a dimension where current models are shown to fail profoundly~\cite{in2025safety}.

The miscalibration in safety systems harms user experience and cripples the model's utility. Research shows that users find simplistic, direct refusals highly frustrating and strongly prefer more nuanced strategies like ``diverting denials" or ``partial compliance," where the model provides helpful, even if non-actionable, information instead of a complete stonewall~\cite{10.1145/3613904.3642135,zheng2025let}. The challenge is not to simply lower the safety threshold, which compromises safety against genuinely malicious attacks~\cite{pmlr-v267-cui25a}, but to build better-calibrated safety systems that can make nuanced risk evaluations about the prompt, context, and user.

\rparagraph{Miscalibration in Reasoning.} 
Models systematically misjudge the quality of their own reasoning. When given the correct first step, they can solve problems they otherwise fail, revealing they possess the capability but cannot accurately assess which reasoning path to take~\cite{jain-etal-2025-first}. This same calibration failure produces ``unfaithful reasoning": models arrive at correct answers via logically flawed rationales, unable to recognize that their stated justification does not match their internal computation~\citep{yee2024dissociation, zhang-etal-2025-roads}. Nor can they identify errors when flawed rationales are provided in-context, confirming they lack the self-monitoring needed to catch mistakes in their own logic~\cite{zhou2024can}.

This miscalibration extends to assessing problem difficulty, where models expend vast computational resources on trivial tasks (e.g., using nearly 1,000 tokens to answer "2+3=?")~\citep{chen2024not}. In agentic settings, this issue leads to a reasoning–action dilemma: higher ``overthinking scores" paradoxically correlate with lower task completion on benchmarks like SWE-bench-Verified, while selecting less-overthought solutions can improve success rates by 30\% and cut compute costs by 43\% \citep{cuadron2025danger}.

At a mechanistic level, this issue is linked to how models handle uncertainty. A well-calibrated reasoner would treat points of low confidence as opportunities for exploration. These decision points often correspond to high-entropy ``forking tokens" (e.g., \texttt{however}, \texttt{alternatively}), which signal a potential shift in the logical chain \citep{cheng2025reasoning}. However, standard reinforcement learning techniques tend to suppress these crucial, often low-probability, ``reasoning sparks." This suppression stifles exploration and funnels the model down a single reasoning path with unwarranted confidence \citep{huang2025low}. This miscalibration makes popular inference-time enhancement strategies less effective and computationally inefficient. For instance, while fine-tuning improves a model's accuracy on a single reasoning attempt (Pass@1), it can cause ``diversity collapse," harming its ability to find the correct answer when given multiple attempts (Pass@K)~\citep{dang2025weight, hamid2025polychromicobjectivesreinforcementlearning}. Self-consistency techniques, which rely on sampling diverse reasoning paths, become ineffective when the model produces only narrow variations of the same flawed logic.

\rparagraph{Calibration Decay in Long Interactions.}
Metacognitive calibration is not static; it erodes over long interactions as models' effective access to conversational history degrades while their expressed confidence remains high. We discuss this phenomenon in detail in Appendix~\ref{app:decay}.

\section{Alternative Views}
\label{sec:alternative-views}

\rparagraph{View 1: These Four Facets Are Separate Problems.}
A natural objection is that the four facets of miscalibration are fundamentally distinct phenomena with different root causes. In this view, probabilistic miscalibration stems from the model mimicking human cognitive biases absorbed from training data. Semantic miscalibration arises from knowledge gaps and the inability to distinguish memorized facts from plausible confabulations. Distributional miscalibration is a direct artifact of preference optimization, which explicitly selects single ``best'' responses. Metacognitive miscalibration reflects a lack of explicit self-monitoring training. Each problem, this view argues, has domain-specific solutions that work in isolation: retrieval-augmented generation for hallucination, higher sampling temperatures for diversity, better safety classifiers for over-refusal. The logical endpoint of this view is architectural separation: train specialized models for each domain (a safety model, a factual QA model, a creative generation model), each optimized for its specific calibration challenge, with a routing system directing queries appropriately.

We argue that these targeted solutions are \textit{patches}, not cures: they address specific failure modes without necessarily correcting the model's miscalibration. Consider the pattern: RAG reduces hallucination by offloading factual recall to an external database, but the model's \textit{internal} confidence remains miscalibrated; it cannot distinguish what it knows from what it retrieved. Temperature scaling increases output diversity, but artificially. It does not teach the model which queries warrant diverse responses versus which have unique correct answers. Safety classifiers reduce over-refusal by adding an external filter, but the model itself still lacks a calibrated sense of risk. 

The specialized models approach merely elevates the problem: the routing system itself must now be calibrated about which model to invoke---a metacognitive calibration challenge. Moreover, it presupposes that queries neatly partition into domains, when real-world prompts often span multiple facets simultaneously (e.g., ``Write a creative but factually accurate story about Marie Curie''). Most critically, each specialized model would still exhibit all facets of miscalibration within its domain. A ``safety specialist'' can still be probabilistically miscalibrated about risk distributions, semantically miscalibrated about what constitutes harm, distributionally collapsing to over-cautious defaults, and metacognitively miscalibrated about when to escalate to human review. In each case, the intervention works around the model's flawed uncertainty representation rather than fixing it. 

Nor do external filters and guardrails, the dominant paradigm in production systems, escape this issue. A hallucination detector must be calibrated about what constitutes a factual error. A safety filter must be calibrated about risk thresholds. Whether the calibration problem lives in the main model or in external components, it remains unsolved. This is why the same failure modes persistently resurface in new forms: conformity \citep{zhu-etal-2025-conformity}, sycophancy \citep{sharma2024towards}, overthinking \citep{chen2024not, cuadron2025danger}, mode collapse \citep{dang2025weight}, lost-in-the-middle \citep{laban2025llms}. Like whack-a-mole, each requires a new bespoke patch because we never addressed the underlying deficit.

Our unified framing does \textit{not} claim that one solution will fix all four facets. Rather, it argues that all four share a common structure: the model's inability to accurately represent its own uncertainty. Sustainable progress requires targeting this root rather than endlessly patching its manifestations.

\rparagraph{View 2: Model Progress Will Naturally Fix Miscalibration.}
This objection holds that miscalibration is just a temporary limitation that will dissolve as models improve. We should focus on advancing the capability frontier; calibration will follow. So far, each generation of LLMs is indeed more capable than the last: better at reasoning, more factually accurate, more instruction-following. From this perspective, calibration will naturally improve as a byproduct of capability gains.

Existing evidence contradicts this view. The relationship between capability and calibration is not monotonic. \citet{cao2025on} show that across models ranging from 0.5B to 70B parameters, larger models are no better calibrated than smaller ones: they accumulate errors during generation at essentially the same rate regardless of scale. This occurs because natural language follows a power law distribution with an exponent close to 1, which theoretically predicts negligible calibration gains from scaling. Larger models can also exhibit less diversity, and alignment stages progressively degrade calibration (see \S\ref{sec:distributional})~\citep{wright2025epistemic,zhang2025noveltybench}. The evidence suggests that current scaling and alignment paradigms do not naturally produce calibrated models; if anything, they may exacerbate certain forms of miscalibration. This is precisely why we argue for calibration as a primary evaluation target: without explicit measurement and optimization pressure, capability gains will continue to mask or worsen calibration failures.

\rparagraph{View 3: Perfect Calibration May Be Undesirable.}
Finally, one might argue that some degree of overconfidence is \textit{functional}. For many applications, maximizing Pass@1 accuracy is the only metric that matters: the user wants the single best answer, not a distribution of hedged possibilities. Furthermore, user studies show that people often prefer confident, direct responses~\citep{10.1145/3613904.3642135}, and hedging can be perceived as evasive or unhelpful.

This argument conflates two distinct concerns. The first is that whether Pass@1 is the right objective is application-dependent. For some use cases, optimizing single-shot accuracy may be appropriate. But this use is precisely why calibration matters: a well-calibrated model enables \textit{informed} deployment decisions. System designers need the confidence signal to decide when to present a single answer versus multiple options, when to route to human review, or when to invoke external tools. If the model cannot produce reliable uncertainty estimates, these choices become impossible. The second concern is how uncertainty should be presented to users, which is a system-level design decision. It presupposes we have access to meaningful confidence signals in the first place. We must solve the ML problem (extracting accurate confidence) before the interface design problem (presenting it effectively) can even be posed. Our position is not that all deployments should display uncertainty, but that all models should \textit{have} calibrated uncertainty available as a primitive for downstream system design.

\section{From Capabilities to Reliability: A Research Agenda}
\label{sec:outlook}

Understanding these failures as facets of miscalibration
requires changes to how LLMs are evaluated, trained, and
deployed. We identify five areas where current practices
fall short and propose concrete directions for improvement.

\paragraph{Measure: A Call for Calibrated Benchmarking}
For model developers, calibration was once an important metric; the GPT-4 report, for instance, featured detailed calibration curves~\cite{achiam2023gpt}. Yet, as the field has raced toward maximizing aggregate benchmarks, this practice has disappeared. At the time of writing, no subsequent major model releases have explicitly evaluated calibration (see Appendix~\ref{app:model-survey}). We face a transparency crisis: while community leaderboards meticulously track accuracy, they mostly ignore calibration. 

This creates a blind spot regarding model reliability. A model that scores 90\% on a benchmark while being equally confident in its correct and incorrect answers is fundamentally different from one that scores 90\% and appropriately expresses uncertainty. Without measuring the alignment between confidence and correctness, we cannot determine whether a model genuinely knows when it has the right answer or is merely guessing with unwarranted certainty. We therefore need to evaluate whether confidence reflects empirical accuracy. This matters for deployment, where confidence estimates can guide decisions about when to invoke external tools, defer to human review, or present multiple answers.

Breaking this cycle requires action from benchmark developers and the research community alike. Every benchmark evaluation should report calibration metrics (e.g., Expected Calibration Error, Brier scores) alongside accuracy. For generative tasks, diversity metrics such as Self-BLEU or unique $n$-gram ratios should become standard. Conference reviewers should ask of empirical papers: ``\textit{How does this model's calibration compare to baselines?}"  Beyond evaluation, reliable uncertainty estimates can also inform model development, from selecting informative training examples~\citep{pmlr-v235-dwaracherla24a} to improving reward-model supervision and identifying when stronger LLM judges are needed~\citep{pmlr-v235-lee24z,xu2025neurips-ask}.

Looking ahead, the community needs a \textbf{holistic calibration benchmark suite} that evaluates the four facets of miscalibration simultaneously. Such a benchmark would not merely check if an answer is right, but would quantify the alignment of confidence scores with correctness (Semantic), measure whether models can enact specified probability distributions (Probabilistic), assess output diversity against human baselines (Distributional), and evaluate the utility of metacognitive decisions like tool use and abstention (Metacognitive). Recent efforts like HELM~\citep{liang2022holistic} and the Humanity's Last Exam leaderboard~\citep{scale_HLE_leaderboard} point in this direction, but calibration-aware evaluation must become the norm, not the exception. As LLMs become embedded in high-stakes decision-making tasks such as medical diagnosis and legal research, miscalibration poses immediate, measurable harm that our current evaluation practices actively obscure. Agentic systems require particular attention, as miscalibration can compound across sequential decisions. Agent benchmarks should report outcome-level calibration alongside task success, while further research is needed to understand how calibration should be defined and evaluated across intermediate decisions
and entire execution trajectories~\citep{subramani-etal-2025-mice,xuan-etal-2026-confidence,pmlr-v306-zhang26gy}.

\paragraph{Train: The Alignment-Calibration Trade-off}
A critical open question is whether current alignment paradigms are fundamentally incompatible with calibration. Base models are often well-calibrated: their token probabilities reflect genuine uncertainty about what comes next~\citep{achiam2023gpt}. However, alignment techniques like RLHF~\citep{ouyang2022training} typically optimize for a point estimate of goodness, training the model to converge on a single, highly-rated response. This creates an alignment-calibration trade-off: the process of aligning a model to be more helpful and safe systematically degrades its calibration~\citep{hu_2025_modelmerging}.

If post-training is part of the problem, model developers must rethink current alignment paradigms. One path forward is to move beyond point-estimate optimization entirely and explore new frontiers like distributional RL or preference models that reward a portfolio of diverse outputs \citep{zhou2025qsharp, lanchantin2025diverse,puri2026escaping}. Another is to preserve the capabilities of base models through collaborative inference~\citep{feng2025dontthrowawaypretrained} or model merging~\citep{hu_2025_modelmerging}. Calibration can also be optimized directly during training, through calibration-aware RL objectives~\citep{damani2025beyond} or models designed for probabilistic decision-making, such as Jev~\citep{almeida2026jev}. At minimum, the practice of publishing calibration curves in technical reports for model releases should be restored, making the calibration cost of alignment visible and accountable.

\paragraph{Deploy: Rethinking the Interaction Design}
Finally, we must address the last mile of miscalibration: the interface. Even a perfectly calibrated model is of limited use if the interaction paradigm acts as an information bottleneck. The dominant chat-based interface forces a lossy compression of the model's high-dimensional, probabilistic belief state into a single, deterministic token sequence. This fosters an \textit{Illusion of Determinism}; users consistently overestimate model accuracy when denied access to internal uncertainty, failing to adjust their trust even when the model errs \citep{steyvers2025large}. Practitioners deploying LLMs should design interaction paradigms that support \textit{Probabilistic Disclosure}, e.g., visualizing confidence intervals or exposing multiple reasoning paths. An alternative approach is linguistic calibration: training models to express uncertainty through natural language (``I'm confident that...'' vs. ``I think maybe...'') rather than numeric probabilities~\citep{mielke-etal-2022-reducing,10.5555/3692070.3692180, zhou-etal-2024-relying}. The goal is not to burden users with probabilities but to ensure that the uncertainty in model outputs reaches the humans who must act on them. 

Importantly, uncertainty need not always be shown
to users. It can instead inform retrieval, clarification,
abstention, or escalation to human review. Model
providers should establish the reliability of their
confidence signals, while downstream developers
determine how these signals are used according to
task-specific requirements and risks. The goal is
not to display uncertainty everywhere, but to make
reliable uncertainty estimates available wherever
they can improve decisions.

\paragraph{Sustain: The Synthetic Data Feedback Loop}
The stakes extend beyond individual deployments to ecosystem-level sustainability. As synthetic data becomes central to training pipelines, miscalibration creates a systemic threat. When models collapse onto narrow, stereotypical outputs, this homogenized text feeds into the next generation of training data, triggering a degenerative feedback loop known as Model Collapse~\citep{shumailov2024ai}. The long-term solution is to fix calibration at the source: ensuring models used for data generation are distributionally calibrated. In the interim, practitioners can explore diversity-preserving patches during data generation, such as persona-based sampling~\citep{hu-collier-2024-quantifying, lambert2025tulu, ge2025scalingsyntheticdatacreation}, or during distillation, preserving multiple valid reasoning chains rather than collapsing to a single canonical path~\citep{guha2025openthoughtsdatarecipesreasoning}. However, these remain symptomatic interventions rather than fundamental fixes. Ultimately, well-calibrated models that maintain diverse, uncertainty-aware outputs are essential not just for today's deployments but for sustaining the quality of tomorrow's training data.

\rparagraph{Generalize: Calibration in Multimodal Systems}
While our analysis focuses on text, the four facets of miscalibration apply across modalities. Text-to-image models exhibit severe stereotypical tendencies even for underspecified prompts (e.g., 98\% of generated ``cookies" sharing the same shape), collapsing the rich distribution of valid visualizations into a narrow default \citep{rassin2024grade}. Multimodal LLMs frequently lack the metacognitive self-awareness to distinguish between visible content and unanswerable queries, confidently answering questions not grounded in the image~\citep{wang-etal-2024-mm}. 

We advocate for a convergence of calibration research across generative modalities. While distinct data modalities have historically necessitated specialized calibration techniques, the emergence of natively multimodal architectures presents a unique opportunity: rather than treating modalities separately, future research should explore methods where the information in one modality (e.g., visual evidence) serves as a calibration signal for another (e.g., textual reasoning).

\section{Conclusion}
Reliable AI systems require models that can accurately assess their own uncertainty. Without this ability, models risk producing confident errors in high-stakes domains and failing to recognize the limits of their competence. This becomes particularly consequential in agentic settings, where models must decide when to act, invoke tools, or seek assistance, and where miscalibrated decisions can propagate through subsequent steps.

Despite substantial advances in model capabilities, calibration has received comparatively little attention
as an evaluation and optimization target. The good news is that base models often exhibit great calibration, which can deteriorate during post-training. We do not need to invent calibration from scratch; we just need to stop destroying it. 

When an LLM failure occurs, the first diagnostic question should be: \emph{what distribution does the model produce, and what distribution should it produce?} We argue that calibration must become a primary evaluation and optimization target alongside accuracy. Achieving this requires rethinking how we train, evaluate, and deploy language models.

\clearpage

\begin{ack}
Tiancheng Hu was supported by the Gates Cambridge Trust.
Caiqi Zhang was supported by an Amazon Studentship.
Dirk Hovy was supported by the European Research Council
under the European Union's Horizon 2020 research and
innovation programme (grant agreement No.~949944,
INTEGRATOR). The authors decalre no competing interest. 
\end{ack}

\bibliographystyle{plainnat}
\bibliography{custom}

\appendix

\section{Limitations}
The boundaries between our four facets of miscalibration are porous by design; many real-world failures exhibit multiple facets simultaneously (e.g., a hallucinated legal citation reflects both semantic and metacognitive miscalibration). Our framework is intended as a diagnostic lens, not a strict taxonomy. Our critique of alignment methods like RLHF highlights their potential to exacerbate certain forms of miscalibration, but we acknowledge their substantial benefits for instruction-following and safety. The precise relationship between alignment objectives, output diversity, and calibration metrics remains an open empirical question requiring systematic investigation beyond the scope of this position paper. Finally, while we argue for the importance of calibration, we do not claim that perfect calibration is always achievable or even desirable across all deployment contexts. The appropriate use of calibrated confidence depends on
the application. Model providers should ensure that
confidence estimates reliably reflect observed outcomes,
while downstream developers must select decision thresholds
based on the costs of errors, abstention, and human review.

\section{Ethical Considerations}
The central theme of this paper, the unreliability of current LLMs, has significant ethical implications. Semantic miscalibration could lead to the confident spread of misinformation. Distributional miscalibration could amplify social stereotypes and marginalize underrepresented groups. Metacognitive miscalibration in safety systems can lead to over-refusal, denying users access to legitimate information, or under-refusal, failing to prevent harm. By providing a unified framework, we hope to encourage research into more robust and reliable models. It is crucial that the ``calibrated benchmarks" we call for are developed with diverse stakeholders to ensure they measure reliability across different cultural contexts and demographic groups.

\section{Calibration: Scope and Relationship to Prior Work}
\label{app:prior-work}

\rparagraph{Calibration Is Not Perfection.}
A calibrated model can still be wrong. Calibration means that 
confidence tracks empirical correctness; it does not imply 
omniscience or zero hallucination~\citep{kalai2024calibrated}. 
A well-calibrated model that is 70\% confident will be wrong 
30\% of the time, but it will be wrong at the rate it predicts. 
The goal is not to eliminate errors but to ensure that a model's 
confidence is informative enough to guide downstream decisions.

\rparagraph{Calibration as Research Topic versus Infrastructure.}
Calibration is not new, and recent work has studied LLM confidence 
estimation, post-hoc calibration, calibration-aware fine-tuning, 
process reward model calibration, and the relationship between 
calibration and hallucination~\citep{shen2024thermometer,
xiao2025restoring,park2025know,kalai2024calibrated,
miao2026hallucination}. Our claim is not that calibration is 
absent from the literature. It is that calibration has not 
become infrastructure: a routine axis of model reporting, 
benchmark design, post-training analysis, deployment routing, 
and interface design. The contribution of this paper is to 
connect failures that are usually studied separately and argue 
that calibration should be treated as the common reliability 
primitive across them.

\rparagraph{Why Four Facets?}
We derived the framework by examining documented LLM
failure modes and identifying, for each, the model signal
being relied upon and the target it should track.
The resulting facets are parallel and overlapping
diagnostic categories rather than a mutually exclusive
partition.

Probabilistic and distributional miscalibration both
concern output distributions, but differ in their
targets. The former evaluates agreement with an
explicitly specified distribution, whereas the latter
concerns a task- or population-level distribution
that must be estimated. Similarly, semantic
miscalibration concerns confidence in individual
claims, while metacognitive miscalibration concerns
the assessment of competence or expected utility
used to select actions.

These distinctions are functional rather than
ontological. A different decomposition is possible,
but the four facets identify different evaluation
targets and suggest different interventions.

\section{Applying the Framework: Premature Closure}
\label{app:diagnostic}

We illustrate the diagnostic use of our framework through
premature closure in clinical reasoning, where models
commit to a diagnosis despite insufficient evidence
~\citep{handler2026quantifyingmitigatingprematureclosure}.

\rparagraph{Identify the signals and targets.}
The relevant signals are the model's confidence in a
diagnosis and its assessment of whether to answer,
clarify, abstain, or escalate. These should track
empirical correctness and the expected utility of
the available actions, respectively.

\rparagraph{Locate the mismatch.}
High confidence in an incorrect diagnosis indicates
semantic miscalibration, while failure to recognize
when clarification or abstention would be preferable
concerns metacognitive miscalibration. These failures
need not co-occur: a model may express low confidence
but still answer because its decision policy favors
responsiveness. If the model also systematically
favors common diagnoses over other plausible
alternatives, distributional miscalibration may
be involved.

\rparagraph{Derive evaluations and interventions.}
The framework suggests evaluating models on paired
answerable and underspecified cases, measuring
both diagnostic confidence and decisions to answer,
clarify, or abstain. Interventions then depend on
the identified mismatch: improving claim-level
calibration, training better clarification and
abstention policies, or increasing coverage of
plausible diagnoses.

\section{Formal Notation and Definitions}
\label{app:notation}

We make the following formal distinctions from ``classical" supervised learning in our four concepts.

\begin{enumerate}
    \item \textbf{Confidence Estimation}:\; Assigns a score $s(x,a) \in [0,1]$ to each question–answer pair $(x,a)$ interpreting the estimated probability that answer $a$ is \emph{correct for} question $x$.
    \item \textbf{Uncertainty Estimation}:\; Quantifies the dispersion of the model’s predictive belief \emph{about possible answers given $x$}, independent of any single answer $a$.
    \item \textbf{Calibration Methods}:\; Procedures that transform raw confidence scores to improve alignment, either post hoc or during training.
    \item \textbf{Calibration Level}:\; A property describing how well (possibly scaled) confidence scores align with empirical correctness.
\end{enumerate}

\rparagraph{Notation.}
Let $\mathcal{X}$ be the input space and $\mathcal{Y}$ the output space (token sequences, classes, spans, etc.).
A model with parameters $\theta$ induces a predictive distribution $q_\theta(a \mid x)$ and a task-specific point prediction $\hat{a}(x) \in \mathcal{Y}$.
For tasks admitting an automatic checker or oracle, define the correctness indicator
\[
  C(x,a) \in \{0,1\},
\]
which returns $1$ iff answer $a$ is correct for input $x$. For a dataset $\mathcal{D}=\{x_i\}_{i=1}^n$, we write $[n]=\{1,\dots,n\}$.

\rparagraph{Why Confidence and Uncertainty Matter.}
Confidence answers \emph{“how likely is this specific answer $a$ to be correct for $x$?”}, guiding acceptance, ranking, and aggregation of model outputs. Uncertainty answers \emph{“how stable are the model’s beliefs over answers for $x$?”}, guiding \emph{when} to trust the model at all—e.g., abstention, human routing, tool use, or clarification—especially in open-ended LLM settings where a calibrated correctness probability may be unavailable.

\rparagraph{Confidence Estimation.}
\label{subsec:confidence}
A confidence estimation method is a function $s:\mathcal{X}\times\mathcal{Y}\!\to\![0,1]$ such that
\[
  s(x,a) \;\approx\; \Pr\!\big(C(x,a)=1 \,\big\vert\, x,a\big).
\]
A common practical choice evaluates the actual predicted answer $\hat{a}(x)$ because downstream decisions act on the single deployed output, and computing correctness for all $a\in\mathcal{Y}$ is impractical:
\[
  s\big(x,\hat{a}(x)\big) \approx \Pr\!\big(C(x,\hat{a}(x))=1 \,\big\vert\, x,\hat{a}(x)\big).
\]

In multiclass classification, $s(x,\hat{a}(x))$ is often the top-label probability $q_\theta(\hat{a}(x)\mid x)$. However, because LLMs generate high-dimensional, free-form text, simple token probabilities often fail to capture true confidence. This necessitates alternative estimation strategies, such as by (i) prompting for verbalized confidence \citep{tian2023just, xiong2024can}, or (ii) sampling multiple candidates and mapping semantic consistency to a correctness probability proxy \citep{manakul2023selfcheckgpt}.

\rparagraph{Uncertainty Estimation}
\label{subsec:uncertainty}
Let $q_\theta(a\mid x)$ be the model’s predictive distribution over answers $a \in \mathcal{Y}$ for input $x$.
\textbf{Uncertainty} is any score that depends only on $q_\theta(\cdot\mid x)$ (and not on a specific answer).
We write
\[
u(x) \;=\; f\!\big(q_\theta(\cdot\mid x)\big),
\]
where $u(x)$ need not be in $[0,1]$. Meanwhile, $u(x)$ is \emph{not} a probability that the model’s chosen answer is correct.
It is used to rank inputs by risk or difficulty (e.g., abstain or route to a human), and is typically evaluated with discrimination metrics such as AUROC/AUPRC rather than calibration error \citep{lin2023generating, kuhnsemantic}.

\rparagraph{Calibration Methods}
\label{subsec:calibration-methods}
Raw confidence scores are often miscalibrated. \textbf{Calibration methods} learn a calibrator $h:[0,1]\!\to\![0,1]$ on held-out data and apply it at test time to achieve better calibration level:
\[
  s'(x,a) = h\!\big(s(x,a)\big)
\]
Post-hoc approaches include temperature scaling and isotonic regression; calibration can also be promoted during training (e.g., via loss design or data augmentation).

\rparagraph{Calibration Level}
\label{subsec:calibration-status}
Calibration level quantifies the alignment between \emph{confidence scores} and \emph{empirical correctness}. Raw scores from confidence estimation are often transformed via calibration methods to optimize this metric. Formally, let $S=s(X,A)$ denote a scalar confidence score, typically evaluated at the model's prediction $A=\hat{a}(X)$. Perfect calibration requires that for all $t \in [0,1]$:
\begin{equation*}
    \Pr(C=1 \mid S=t) \;=\; t.
\end{equation*}

Empirically, this property is visualized using \emph{reliability diagrams} and measured by metrics such as Expected Calibration Error (ECE), Maximum Calibration Error (MCE), and Brier Score. 

\rparagraph{Summary} In this paper, we use \emph{calibration level} to denote calibration as a property, and \emph{calibration method} to denote calibration as a procedure~\citep{zhou2025finallayer}.
Thus, a strong \textit{confidence estimator} combined with an effective \textit{calibration method} is a practical way to improve \textit{calibration level}.
Unless stated otherwise, we use ``calibration'' and ``miscalibration'' to refer to calibration level.

\section{Practical Calibration Measurement}
\label{app:measurement}

\rparagraph{Semantic Equivalence.}
For open-ended generation, distinct strings may express
the same answer. Calibration can therefore be evaluated
over semantic equivalence classes rather than individual
sequences. For longer responses, factual claims or
semantic units provide more suitable evaluation targets
~\citep{kuhnsemantic,zhang-etal-2024-luq}.
However, semantic clustering and claim-level factuality
assessment can themselves introduce errors.

\rparagraph{Confidence Estimation.}
Token probabilities need not provide reliable estimates
of answer correctness after post-training. Alternative
signals include verbalized probabilities, self-evaluation,
semantic consistency across samples, and learned
confidence estimators~\citep{tian2023just,
manakul2023selfcheckgpt,shen2024thermometer}.
These signals can be evaluated against observed
correctness using calibration metrics such as ECE
or the Brier score.

\rparagraph{Large Output Spaces.}
Calibration does not require enumerating every possible
output. Repeated sampling can approximate the relevant
answer distribution, allowing evaluations of semantic
uncertainty, mode coverage, or divergence from a
specified target. Conformal methods provide a
complementary approach through prediction sets or
abstention policies with coverage guarantees under
appropriate assumptions. The appropriate measurement
depends on the task, the definition of correctness,
and the available reference distribution.

\section{Model Release Calibration Survey}
\label{app:model-survey}
At the time of writing, we surveyed the technical reports of every major frontier model 
release following the GPT-4 report~\citep{achiam2023gpt}, which was 
the last to include calibration curves. None of the following 
model cards / reports include calibration metrics: GPT-4V~\citep{openai2023gpt4v},
GPT-4o~\citep{openai2024gpt4o},
o1~\citep{openai2024o1},
o3 and o4-mini~\citep{openai2025o3o4mini},
GPT-5~\citep{OpenAI_GPT5_SystemCard_2025},
GPT-5.5~\citep{openai2026gpt55};
Claude~3~\citep{anthropic2024claude3},
Claude~3.5~\citep{anthropic2024claude35,anthropic2024claude35updated},
Claude~3.7~\citep{anthropic2025claude37},
Claude~4~\citep{anthropic2025claude4},
Claude~4.5~\citep{anthropic2025claudesonnet45,anthropic2025claudeopus45},
Claude~4.6~\citep{anthropic2026claudeopus46,anthropic2026claudesonnet46};
Gemini~1.5~\citep{geminiteam2024gemini15},
Gemini~2.5~\citep{gemini2.5},
Gemini~3 Pro~\citep{googledeepmind2025gemini3pro};
Llama~3~\citep{llama2024llama3},
Llama~4~\citep{meta2025llama4};
DeepSeek-V3~\citep{deepseekai2024deepseekv3},
DeepSeek-R1~\citep{deepseekai2025deepseekr1};
Grok~4 and Grok~4.1~\citep{xai2025grok4,xai2025grok41};
Mistral Large~2~\citep{mistralai2024large2};
Qwen2.5~\citep{qwen2024qwen25},
Qwen3~\citep{qwen2025qwen3};
Command~A~\citep{cohere2025commanda};
and Kimi~K2~\citep{kimi2025kimik2}. While some of these reports evaluate factual accuracy on benchmarks like SimpleQA, none measure the alignment between model confidence and correctness.

\section{Calibration Decay in Long Interactions}
\label{app:decay}
Metacognitive calibration is not static; it erodes over long interactions~\citep{zhang2026multiturn}. A model's expressed confidence remains high even as its effective access to conversational history degrades. This stems partially from attention failure:~\citet{liu-etal-2024-lost} demonstrate a ``lost in the middle" effect, where a model's ability to retrieve information decays based on its position in the context window. In multi-turn dialogue, this manifests as a ``loss-in-middle-turns" effect, where models overweight initial and recent exchanges while forgetting crucial context from the middle~\cite{laban2025llms}. The failure is that models are not calibrated to their attentional blind spots—they operate with certainty misaligned with their decaying ability to access the full history. This leads to ``logical decay"~\cite{lei2025d}, where models confidently contradict facts stated turns prior. It also explains why models that take a wrong turn get ``lost and do not recover"~\cite{laban2025llms}: miscalibrated self-assessment prevents them from recognizing their own error.

\end{document}